\documentclass{article}

\PassOptionsToPackage{numbers, compress}{natbib}
    \usepackage[final]{neurips_2021}

\usepackage[utf8]{inputenc} % allow utf-8 input
\usepackage[T1]{fontenc}    % use 8-bit T1 fonts
\usepackage{hyperref}       % hyperlinks
\usepackage{url}            % simple URL typesetting
\usepackage{booktabs}       % professional-quality tables
\usepackage{amsfonts}       % blackboard math symbols
\usepackage{nicefrac}       % compact symbols for 1/2, etc.
\usepackage{microtype}      % microtypography
\usepackage{xcolor}         % colors
\usepackage{graphicx}

\title{Solving Traffic4Cast Competition with U-Net and Temporal Domain Adaptation}

\author{%
  Vsevolod Konyakhin \\
  ITMO University\\
  Saint Petersburg, Russia \\
  \texttt{sevakonyakhin@gmail.com} \\
   \And
   Nina Lukashina \\
   JetBrains Research \\
   Saint Petersburg, Russia \\
   \texttt{nina.lukashina@jetbrains.com} \\
   \AND
   Aleksei Shpilman \\
   JetBrains Research \\
   HSE University \\
   Saint Petersburg, Russia \\
   \texttt{alexey@shpilman.com} \\
}

\begin{document}

\maketitle

\begin{abstract}
  In this technical report, we present our solution to the Traffic4Cast 2021 Core Challenge, in which participants were asked to develop algorithms for predicting a traffic state 60 minutes ahead, based on the information from the previous hour, in 4 different cities. In contrast to the previously held competitions, this year's challenge focuses on the temporal domain shift in traffic due to the COVID-19 pandemic. Following the past success of U-Net, we utilize it for predicting future traffic maps. Additionally, we explore the usage of pre-trained encoders such as DenseNet and EfficientNet and employ multiple domain adaptation techniques to fight the domain shift. Our solution has ranked third in the final competition. The code is available at \url{https://github.com/jbr-ai-labs/traffic4cast-2021}.
\end{abstract}

\section{Introduction}
% Traffic state prediction is an active area of research. Two more sentences.

The Traffic4Cast Competition, organized over the last three years by the Institute of Advanced Research in Artificial Intelligence (IARAI) and the data-provider company, HERE technologies, \footnote{\url{https://www.iarai.ac.at/traffic4cast/} and \url{https://www.here.com/}} challenges its participants to design algorithms that would predict future traffic states based on the prior traffic information. Specifically, the Traffic4Cast 2019 Challenge \cite{kreil2020surprising} focuses on forecasting the traffic state 15 minutes ahead based on the dynamic information from the previous hour, while the prediction window is increased to up to an hour in the Traffic4Cast 2020 Challenge \cite{kopp2021traffic4cast}.

This year's edition of the competition, Traffic4Cast 2021, has introduced two competition tracks: 
\begin{itemize}
\item Core Challenge, in which participants are tasked to fight a temporal domain shift that appears due to the COVID-19 pandemic.
\item Extended Challenge, in which models are tested over 2 entirely unseen cities, thus introducing a spatiotemporal domain shift in the traffic data. 
\end{itemize}

Our work focuses on the Core Challenge, for which the competition organizers have provided 6 months of training data from 2019. They have tested the submissions on 100 one-hour time slots from 2020 in four diverse cities — Melbourne, Berlin, Istanbul, Chicago. For each one-hour time slot, the designed algorithms have been challenged to predict traffic conditions for the next six time steps into the future: 5, 10, 15, 30, 45, and 60 minutes. The main difficulty of this challenge is the temporal domain shift, which appears due to the COVID-19 pandemic in 2020 that had inevitably reduced the traffic flow volume due to mobility limitations in order to abate the spread of the virus. For participants, this has imposed a problem where the train and test distributions differ significantly. 

Our contributions are as follows: 
\begin{itemize}
    \item Inspired by the success of U-Net \cite{ronneberger2015u} in the previous competitions \cite{choi2020utilizing, xu2020good, choi2019traffic, martin2019traffic4cast,liu2019building}, we build and train four independent U-Net models for each of the core cities, which serve as a \textit{simple, yet strong} baseline that turned out to be hard to beat.
    \item In attempt to further improve the score, we have conducted experiments employing DenseNet \cite{huang2017densely} and EfficientNet \cite{tan2019efficientnet}, convolutional networks pre-trained on ImageNet \cite{5206848}, as part of the encoder in U-Net \cite{ronneberger2015u}. 
    \item We further have tried a set of domain adaptation techniques, such as pseudo-labeling \cite{jaiswal2019semisupervised} and heuristic map pre- and post-processing to fight the temporal domain shift between the training and test data.
    \item Finally, we have used an ensembling method to combine the networks' predictions, utilizing  the proposed domain adaptation technique, to create the best scoring submission and have ranked \textbf{third} in the competition. 
\end{itemize}

\section{Data Preprocessing}
\label{data}

The data had been collected from 4 cities: Berlin, Istanbul, Melbourne, Chicago. The competition organizers represent each city as a grid map of $495 \times 436$ pixels, where each cell aggregates the GPS measurements in a region of approximately $100 \times 100$ meters size and 5-minute time bins. Then, the dynamic traffic state at each 5-minute time bin is encoded into a grid map with 8 channels that represent traffic volume and speed in four directions (NW, NE, SW, SE). The "pixel values" of volume and speed are scaled to values from 0 to 255 with a min-max scaler and then rounded to their integer values.

Resulting input and output shapes are $T_{in} \times C \times H \times W $ and $T_{out} \times C \times H \times W $ , where $T_{in} = 12$, $T_{out} = 6$, $C = 8$, $H = 495$, $W = 436$, although $H$ and $W$ might differ with different models, in case we pad the input image to match the working resolution of a specific model. The input represents the dynamic traffic state map over 60 minutes window, and the output includes timestamps 5, 10, 15, 30, 45 and 60 minutes after. We further reshape the input and output into three-dimensional tensors to efficiently use image-to-image models, so the resulting shapes are $(T_{in} \times C) \times H \times W $ for input and $(T_{out} \times C) \times H \times W $ for output, respectively. 

As we only have 6 months of labeled data from 2019, we further randomly split the data into 4 folds, fixing one of the folds as a validation set. 

\begin{figure}
  \centering
    \includegraphics[width=\linewidth]{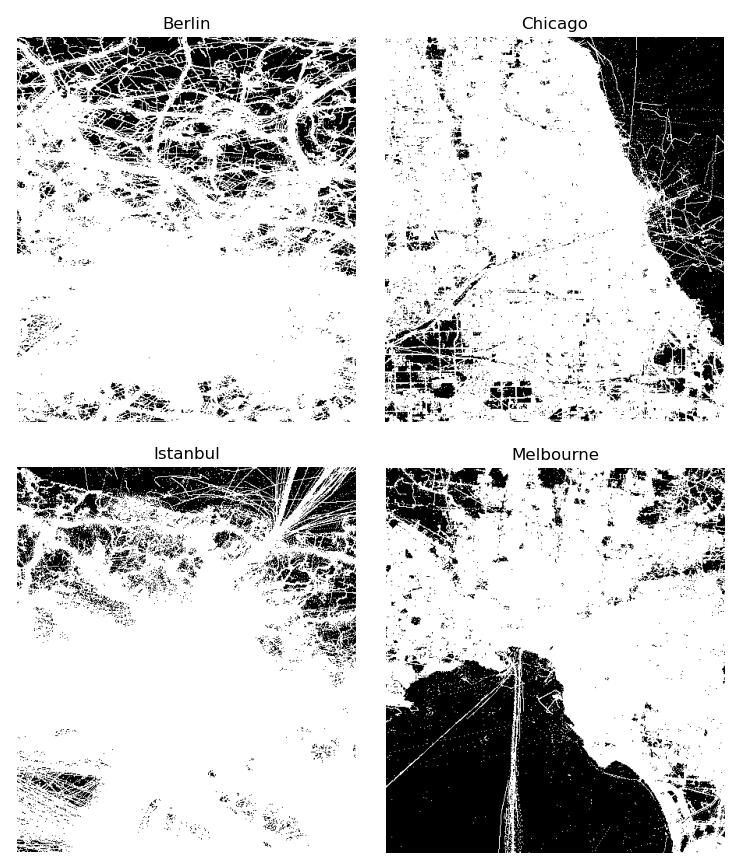}
  \caption{Generated masks of the road map from train and test data, for Berlin, Chicago, Istanbul, and Melbourne.}
\label{fig:maps}
\end{figure}

\paragraph{Static roadmap generation}
The predicted traffic maps are sparse, meaning that there are pixels that never encode any traffic since there is simply no road in that region. While we expect that a large part of the road network knowledge will be inferred into the trained networks, we find that applying post-processing by simply multiplying a two-dimensional binary mask of a road graph with the network predictions yields better results. For this 2D mask, we experiment with the low-resolution static maps, provided by the competition organizers, alongside with the masks that were generated from training data the following way: we set a pixel value to $1$ if the traffic volume or speed value has at least once been more than zero among the training samples, and $0$ otherwise. We find that adding pixel values from 2020 test data further boosts the performance. The obtained roadmaps are shown in Figure \ref{fig:maps}.

\section{Methods}

\subsection{U-Net}
As a strong baseline solution, we utilize a vanilla U-Net network \cite{ronneberger2015u}, with 5 downsampling blocks and batch normalization \cite{ioffe2015batch} following each convolutional block with ReLU activation \cite{nair2010rectified}. The U-Net network takes an image with $T_{in} \times C = 12 \times 8 = 96$ channels and outputs an image with $T_{out} \times C = 6 \times 8 = 48$ channels. We pad the input maps to a spatial resolution of $496 \times 448$ pixels to better handle the downsampling. 

\paragraph{Baseline training details} 
We train an independent model for each city, with the \emph{mean squared error} (MSE) chosen as the loss function to match the evaluation metric. PyTorch framework \cite{NEURIPS2019_9015} is used for all of our experiments. We employ Adam \cite{kingma2014adam} as the optimizer with the learning rate of $3e^{-4}$ and the batch size of $8$. We also employ a warm-up technique by linearly scaling the learning rate from $0$ to $3e^{-4}$ for the first 1k optimization steps. All our training experiments are conducted with the Mixed Precision \cite{micikevicius2018mixed} in order to decrease training time. 

\paragraph{Results}
The U-Net performance on the validation set of the traffic data from 2019 per city is reported in Table~\ref{val2019-table}. We furthermore evaluate on the organizer's test set and report the scores from the leaderboard for the post-processing masks discussed in Section \ref{data}. Table~\ref{test2020-table} demonstrates that our proposed roadmap generation method yields the best result on the competition leaderboard. This post-processing technique is later used by default for all submissions. 

It is worth noting that these 4 standalone U-Net models with the roadmap post-processing have been sufficient to rank third in this year's Core Challenge, meaning that carefully tuned U-Net models provide a simple, yet strong solution for the traffic forecasting task that is robust to the temporal domain shift. 

\begin{table}
  \caption{Models' performance over the fixed validation set from 2019 data. Mean Squared Error, the lower the better.}
  \label{val2019-table}
  \centering
  \begin{tabular}{llllll}
    \toprule
    \multicolumn{5}{r}{Mean Squared Error}                   \\
    \cmidrule(r){3-6}
    Model    & \#Params & Melbourne      & Berlin     &    Istanbul  &   Chicago \\
    \midrule
    U-Net & 31M &  $26.759$ & $80.903$ & $56.623$ & $42.663$ \\
    EfficientNet-B5 U-Net & 30M & $26.013$ & $80.351$ & $\mathbf{55.792}$ & $41.643$ \\
    DenseNet201 U-Net & 31M & $\mathbf{25.739}$ & $\mathbf{78.430}$ & - & $\mathbf{41.158}$ \\
    \bottomrule
  \end{tabular}
\end{table}

\begin{table}
  \caption{U-Net performance over the test set from 2020 data. Mean Squared Error, the lower the better.}
  \label{test2020-table}
  \centering
  \begin{tabular}{ll}
    \toprule
    % \multicolumn{5}{r}{Mean Squared Error}                   \\
    % \cmidrule(r){2}
    Model  & MSE $\downarrow$ \\
    \midrule
    U-Net + static city mask from competition organizers & $49.73289$ \\
    U-Net & $49.69502$ \\
    U-Net + generated static city mask from 2019 train data & $49.69491$ \\
    U-Net + generated static city mask from 2019 train data and 2020 test data & $\textbf{49.69488}$ \\
    \bottomrule
  \end{tabular}
\end{table}

\subsection{Pre-trained Encoders}
Inspired by the success of employing convolutional networks, pre-trained on ImageNet \cite{5206848}, as encoders into U-Net \cite{ronneberger2015u} for a task of semantic segmentation \cite{iglovikov2018ternausnet}, we try two different convolutional networks: DenseNet201 \cite{huang2017densely} and EfficientNet-B5 \cite{tan2019efficientnet}. 

\paragraph{EfficientNet}
EfficientNet \cite{tan2019efficientnet} has been widely used in many vision tasks, such as detection \cite{tan2020efficientdet}, segmentation \cite{Huynh2020AUW}, and others, and has proved to be powerful. Thus, we employ EfficientNet as the encoder part, and keep the decoder from the original U-Net architecture, changing the number of input channels in the decoder side to match the feature maps produced by the EfficientNet. We employ the EfficientNet-B5, pre-trained on ImageNet, so the resulting network achieves almost the same number of parameters as the baseline U-Net, as shown in Table~\ref{val2019-table}. The depth of the U-Net is increased by 1 to match the number of feature maps obtained from the EfficientNet backbone. 

\paragraph{DenseNet}
Following the insight from the last two competitions' winner \cite{choi2019traffic, choi2020utilizing}, we utilize the densely connected convolutional layers by employing DenseNet201 \cite{huang2017densely} pre-trained on ImageNet. Same as in EfficientNet, the decoder part is borrowed from the U-Net; however, since we omit the last dense block of DenseNet to keep the consistency across the size of the models, the resulting network's depth is not increased compared to the baseline U-Net. 

\paragraph{Additional training details}
The EfficientNet- and DenseNet U-Nets are trained with the same parameters as the baseline U-Net, with the two differences: we increase the batch size from 8 to 16 to stabilize the training process, and also increase the padding for the input images, so the traffic maps have a spatial resolution of $512 \times 448$ pixels. 

\begin{table}
  \caption{Comparison of the baseline U-Net and pre-trained encoder U-Nets over the test set from 2020 data. Mean Squared Error, the lower the better.}
  \label{encoders-test2020-table}
  \centering
  \begin{tabular}{ll}
    \toprule
    % \multicolumn{5}{r}{Mean Squared Error}                   \\
    % \cmidrule(r){2}
    Model  & MSE $\downarrow$ \\
    \midrule
    EfficientNet-B5 U-Net + generated static city mask from 2019 and 2020 data & $50.57098$ \\
    DenseNet201 U-Net + generated static city mask from 2019   and 2020 data \ & $49.80787$ \\
    U-Net + generated static city mask from 2019 and 2020 data & $\textbf{49.69488}$ \\
    \bottomrule
  \end{tabular}
\end{table}

\paragraph{Results}
The performance of the two networks over the 2019 validation set can be seen in Table~\ref{val2019-table}. The EfficientNet-B5 U-Net achieves a smaller MSE over all the core cities, when compared to the baseline U-Net, and the DenseNet-201 U-Net performs even better than the EfficientNet-B5 U-Net. Unfortunately, the training process of DenseNet201 U-Net on Istanbul suffered numerical instabilities, so we don't use the DenseNet201 U-Net for this city thereafter.

However, these U-Net models with the pre-trained encoders, although showing better results during the validation on 2019 data, were not able to beat the baseline U-Net on the competition leaderboard (Table~\ref{encoders-test2020-table}).\footnote{As there is no DenseNet U-Net model trained for Istanbul, we use predictions from the Istanbul baseline U-Net model}. This observation guided our further research direction into finding ways of bridging the gap between the 2019 training and 2020 testing data, which is further described in Section \ref{domainadapt}.

\subsection{Domain Adaptation}
\label{domainadapt}
Domain Adaptation is a common technique for tackling the problem when the model was trained on one domain, and tested on a different, but close target domain.  As in the Core Challenge, the 2020 test data differs significantly due to a temporal domain shift, we study two Domain Adaptation strategies applicable to the 2021 Traffic4Cast Core Challenge. 

\paragraph{Pseudo-labeling}
One of the studied domain adaptation approaches is pseudo-labeling \cite{jaiswal2019semisupervised}. The idea is the following: for the unlabeled, out-of-domain data, a model predicts new labels for these samples, thus generating a new labeled data set which is later used for training the model. This procedure might be applied for several rounds. When tried for the baseline U-Net and the 2020 test data, however, our models seem to have degraded their performance on the competition leaderboard. We explain that behavior by the fact that for the 2020 test data, the temporal domain shift is also present in the output maps. 

\paragraph{Heuristically bridging the 2019 and 2020 data}
Instead of modifying the model itself, we focus on finding a way to make the 2020 testing data closer to the 2019 data, which the above-mentioned models have been trained on. We try to find a deterministic inverse transformation that would make the 2020 traffic map closer to the 2019 map, use the transformed maps to generate the model predictions, and apply the inverse transformation on predictions to return to the 2020 test data domain.

Let us denote the traffic map at a certain time step $t$ as $M_t \in \mathbb{R}^{H \times W \times C}$, where $H = 495$, $W = 436$, $C = 8$. Then, for all the time steps from 2019 $t \in T_{2019}$ and the time steps from 2020 $t \in T_{2020}$ we calculate the mean traffic map per year the following way:

$$ M_{2019} = \frac{1}{\# T_{2019} } \sum\nolimits_{t \in T_{2019}} M_t$$
$$ M_{2020} = \frac{1}{\# T_{2020} } \sum\nolimits_{t \in T_{2020}} M_t$$

where the summation symbol denotes element-wise sum. With the mean traffic maps $M_{2019}$ and $M_{2020}$ for 2019 training and 2020 test data, we then calculate per-pixel and per-channel relation $\lambda$ of the mean traffic map $M_{2019}$ to the mean traffic map $M_{2020}$:
$$ \lambda = M_{2019} \oslash M_{2020} $$
where $\oslash$ denotes element-wise division. In case the element of $M_{2020}$ is $0$, we set the corresponding element in $\lambda$ to $1$. Furthermore, based on the fact that the  traffic  volume could not have increased in 2020, we change the outlier elements in $\lambda$ which are less than $1$ to $1$. 

We then multiply the input traffic maps from 2020 test data by $\lambda$, so the traffic maps' distribution becomes closer to the 2019 training data, run the model inference, and then multiply the predicted traffic maps by $\frac{1}{\lambda}$\footnote{This notation denotes that for each element $x$ in $\lambda$, the resulting corresponding element would be $1/x$. In case $x = 0$, the value is set to $1$}, to bring the distribution back to the 2020 test data. 

Using this \textit{Temporal Domain Adaptation} later would result in our best submission on the competition leaderboard, as described in Section \ref{ensembling}.

\subsection{Ensembling}
\label{ensembling}
 Having had multiple models and domain adaptation strategies by the end of the competition, we have ultimately experimented with ensembling different models and strategies together. As discussed in Section \ref{data}, for all the predictions we have utilized our proposed generated static city mask as the post-processing function. We experimented with two aggregating functions, namely mean and median ensembling, and chose the mean ensembling as it had performed slightly better. 
 
 Our three final ensembling submissions are the following:
 \begin{itemize}
     \item The mean ensemble of the predictions from the Baseline U-Net, EfficientNet-B5 U-Net, and DenseNet201 U-Net.
     \item The mean ensemble of the predictions from the Baseline U-Net, EfficientNet-B5 U-Net, and DenseNet201 U-Net with the  \textit{Temporal Domain Adaptation} (TDA) from Section \ref{domainadapt} 
     \item The two above submission combined
 \end{itemize}
 
 We report the obtained errors on the competition leaderboard in Table \ref{final-test2020-table}. The ensemble of the Baseline U-Net, EfficientNet-B5 U-Net, and DenseNet201 U-Net models outperforms the standalone Baseline U-Net models; what's more, when combining this ensemble with the predictions generated from the same models, but using the Temporal Domain Adaptation strategy, the error is further decreased to $\mathbf{49.37906}$, which makes it our \textbf{best} submission. 

\begin{table}
  \caption{Ensembled models performance over the test set from 2020 data.  Mean Squared Error, the lower the better.}
  \label{final-test2020-table}
  \centering
  \begin{tabular}{ll}
    \toprule
    % \multicolumn{5}{r}{Mean Squared Error}                   \\
    % \cmidrule(r){2}
    Ensemble  & MSE $\downarrow$ \\
    \midrule
    Baseline U-Net, EfficientNet-B5 U-Net, DenseNet201 U-Net with TDA & $49.87972$  \\
    Baseline U-Net, EfficientNet-B5 U-Net, DenseNet201 U-Net & $49.45481$ \\
    The two of the above combined & $\mathbf{49.37906}$ \\ 
    \bottomrule
  \end{tabular}
\end{table}

\section{Conclusion}
Our Traffic4Cast 2021 Core Challenge solution yet again shows that U-Net \cite{ronneberger2015u} is a strong baseline for the traffic prediction task. We have studied further into employing convolutional encoders, pre-trained on ImageNet  \cite{5206848}, to improve the baseline U-Net model; however, the obtained models were less robust to the temporal domain shift. In order to fight the latter, we have applied an ensembling method with the introduced \textit{Temporal Domain Adaptation} that aims to bridge the gap between the 2019 and 2020 data. The method has ranked third on the final leaderboard.

\bibliographystyle{unsrt}

\bibliography{biblio}

\end{document}